\documentclass[letterpaper, 10 pt, conference]{ieeeconf}  

\IEEEoverridecommandlockouts             

\usepackage{graphicx}
\usepackage{subfigure}
\usepackage[ruled,linesnumbered]{algorithm2e}
\usepackage{algpseudocode}
\usepackage{multirow}
\usepackage{multicol}
\usepackage{makecell}
\usepackage{textcomp}
\usepackage{stfloats}
\usepackage{cite}
\usepackage{amsfonts,amssymb}
\usepackage{amsmath}
\usepackage{colortbl}
\usepackage{xcolor}
\usepackage[colorlinks,
            linkcolor=red,
            anchorcolor=green,
            citecolor=blue]{hyperref}

\title{\LARGE \bf
HEROS: Hierarchical Exploration with Online Subregion Updating for 3D Environment Coverage
}

\author{Shijun Long\textsuperscript{1}, Ying Li\textsuperscript{1}, Chenming Wu\textsuperscript{2}, Bin Xu\textsuperscript{1}, and Wei Fan\textsuperscript{1} 
\thanks{This work was supported in part by the National Natural Science Foundation of China under Grant 52102449, in part by the Natural Science Foundation of Chongqing under Grant CSTB2023NSCQ-MSX0550, and in part by the Beijing Institute of Technology Research Fund Program for Young Scholars and S\&T Program of Hebei under Grant 21567606H.}
\thanks{S. Long, Y. Li, B. Xu, and F. Wei are with the School of Mechanical Engineering, Beijing Institute of Technology, Beijing, China. {\tt\small \{sj\_long, ying.li, bitxubin, fanweixx\}@bit.edu.cn}}%
\thanks{C. Wu is with RAL, Baidu Research. {\tt\small wuchenming@baidu.com}}
}

\begin{document}

\maketitle
\thispagestyle{empty}
\pagestyle{empty}

\begin{abstract}
We present an autonomous exploration system for efficient coverage of unknown environments. First, a rapid environment preprocessing method is introduced to provide environmental information for subsequent exploration planning. The exploration space is then divided into multiple subregion cells with varying levels of detail, capable of online decomposition and updating to adaptively represent dynamic unknown regions. Finally, a hierarchical planning strategy treats subregions as basic units, computing an efficient global coverage path. The local path is refined under the guidance of the global path to sequentially visit the viewpoint set, providing an executable trajectory for the robot. This coarse-to-fine planning reduces complexity while enhancing exploration efficiency. Compared with state-of-the-art methods in benchmark environments, the proposed approach achieves superior exploration efficiency with lower computational costs.
\end{abstract}

\section{Introduction}
Autonomous exploration using ground mobile robots has received considerable attention in recent years for its wide applications, including searching and rescuing, environmental monitoring, and other tasks. The ability of robots to navigate and explore unknown environments without human intervention enhances operational efficiency and ensures personnel safety in hazardous conditions. Although several autonomous exploration methods have been proposed over the decades \cite{frontier_based, nbvp, rapid_exploration}, the task remains challenging in large-scale and topologically complex scenarios. In such scenarios, the problem becomes computationally complex, and the robot needs to navigate between areas in a limited time while avoiding redundant
exploration.

The mainstream exploration frameworks mainly include frontier-based and sampling-based methods. In autonomous exploration, the frontier is defined as the boundary between free and unknown areas \cite{frontier_based}. These two approaches guide the robot to cover the environment by continuously searching for frontiers in the map or randomly sampling viewpoints and evaluating their potential revenue. However, several limitations still hinder the development of autonomous exploration:

\textit{High Computational Cost}. Both frontier detection and revenue evaluation are computationally expensive, requiring frequent checks of each voxel's occupancy status or performing ray tracing \cite{gbp, nbvp, dsvp}. This presents a substantial challenge for ground-based platforms equipped with edge computing devices to meet real-time requirements.

\textit{Low Efficiency}. Most methods \cite{nbvp, ufoexplorer, frontier_based} employ a greedy strategy of direct traveling to the frontier or viewpoint with the highest gain, disregarding the global coverage path. However, it is crucial to consider the potential impact on future exploration planning when selecting the next target, to avoid inefficiencies caused by shortsighted decisions. While some approaches \cite{two-stage}, \cite{distance} employ the Travelling Salesman Problem (TSP) to obtain the global coverage path, as the number of viewpoints increases, the TSP becomes larger and heavier to solve due to its NP-Complete natural \cite{dsvp}.

\begin{figure}[!t]\centering
    \includegraphics[width=7cm]{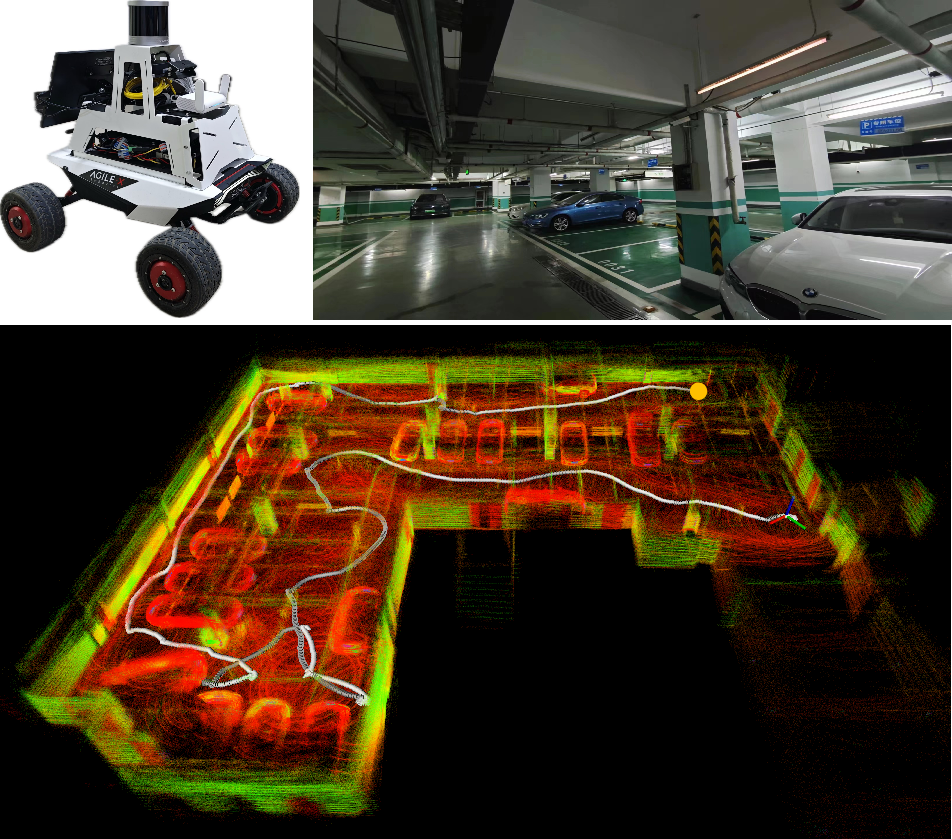} 
    \caption{The real-world experiment conducted in an underground garage using real ground platforms. The yellow dot denotes the start position, and the white line indicates the robot's trajectory. Video of the experiment is available at: \href{https://youtu.be/5shy7c1Faf4}{https://youtu.be/5shy7c1Faf4}.}
\label{real_map}
\end{figure}

In our previous work \cite{long2024hphs}, the hierarchical planning strategy demonstrates significant advantages in autonomous exploration. However, during space division, all subregions use the same resolution, which limits their ability to adapt to variations in the scale and structure of unknown areas, thereby reducing the capability for refined representation. Furthermore, the local planning process relies on a reward-based strategy, which results in shortsightedness, reduced efficiency in large-scale environments, and challenges in ensuring global optimization in complex scenarios.

In this work, we propose an efficient hierarchical exploration planner for autonomous exploration in complex and large-scale environments. First, an incremental environmental preprocessing method is introduced to generate a set of high-quality viewpoints as candidate target points. To better capture the dynamics of unknown areas during exploration, a variable-resolution regional division approach is applied, enabling subregion cells to be further subdivided and adjusted as the unknown areas are transformed into known areas during exploration. Additionally, the Subregion Information Structure (SIS) is designed to record essential information within the subregion space, facilitating exploration planning. Finally, subregions are employed as the basic planning units, and a hierarchical planning strategy is developed for efficient exploration. The global coverage path determines the shortest travel route that sequentially visits each subregion, while the local path refines viewpoint access under the guidance of the global path. This process is continuously executed until the entire environment is fully explored.

 The main contributions of this paper are as follows:

\begin{itemize}

\item A subregion-based regional division method with a hierarchical structure is proposed, enabling adaptive decomposition of subregions to effectively represent the evolving characteristics of unknown regions.

\item A hierarchical planning strategy is adopted, achieving a significant balance between coverage exploration efficiency and computational complexity. 

\item Extensive simulation and real-world experiments are conducted to verify the theoretical and practical feasibility of the proposed method. Our approach demonstrates the fastest exploration with minimal computational resource consumption across diverse scenarios.
\end{itemize}

The source code will be released for further research$\footnote{Available at {\tt\small https://github.com/bit-lsj/HEROS.git}}$.

\section{Related Works}
The field of autonomous robotic exploration has seen significant advancements in recent years, with a variety of approaches being proposed to tackle the challenges of unknown and complex environments. Early work in this domain focused on heuristic-based methods, such as frontier-based exploration \cite{frontier_based, rapid_exploration, fast_mav}. This type of method typically selects the nearest frontier or the frontier with the highest information gain as the next target. They have worked in simple environments but often fall short in complex large-scale scenes due to their greediness. 
Besides, these methods still require a significant amount of frontier detection and clustering, making it difficult to meet real-time requirements for unmanned platforms. \cite{fuel} introduces an incremental frontier detection method and performs global coverage path planning after clustering frontiers. Nevertheless, in large-scale environments, the algorithm's runtime speed is severely slowed down due to a large number of clusters and solving TSP. \cite{fael}  greatly improves planning efficiency by focusing on a specific range of viewpoints for exploration planning. However, it lacks global path guidance and can still lead to inefficient exploration in some cases.

The sampling-based method evaluates sampled viewpoints using a utility function and selects the viewpoint or branch with the highest gain to observe the unknown region. The idea of the next best views \cite{nbv} is first introduced into autonomous exploration by \cite{nbvp}. This method utilizes rapidly random trees (RRT) to explore 3D spaces, selecting the branch with the highest gain for execution. This method has been continuously improved by \cite{gbp}, \cite{dsvp}, but they all suffer from the trouble of expensive computational costs. The majority of computational resources for these methods are consumed in evaluating the gains of viewpoints or branches, which requires estimating the quantity of a large number of unknown voxels.

Several methods \cite{cure}, \cite{racer}, \cite{tdle}, \cite{tare} are proposed for the concepts of regional division and hierarchical planning. However, these methods are either designed for task allocation in multi-robot collaborative exploration or utilize fixed-sized regions that cannot adapt to the scale and structural changes of the unknown region. In \cite{tdle}, the global path is introduced to provide a more comprehensive perspective for exploration. However, the greedy strategy is still adopted in the local planning stage, which can still lead to short-sighted behavior and thus reduce exploration efficiency. The concept of region hierarchical division in this paper is inspired by \cite{racer}, which utilizes hgrid space for muti-UAV interaction and task assignment. Our approach of the regional division is aimed at serving the global coverage path planning, which is part of the hierarchical planning.

Learning-based techniques have also emerged as promising solutions for autonomous exploration \cite{learning_1}, \cite{learning_2}. These methods leverage data-driven models to predict optimal exploration strategies but often lack robustness and adaptability to unfamiliar scenarios. This leads to performance deterioration in complex and unfamiliar environments.

\begin{figure}[!t]\centering
\includegraphics[width=6.3cm]{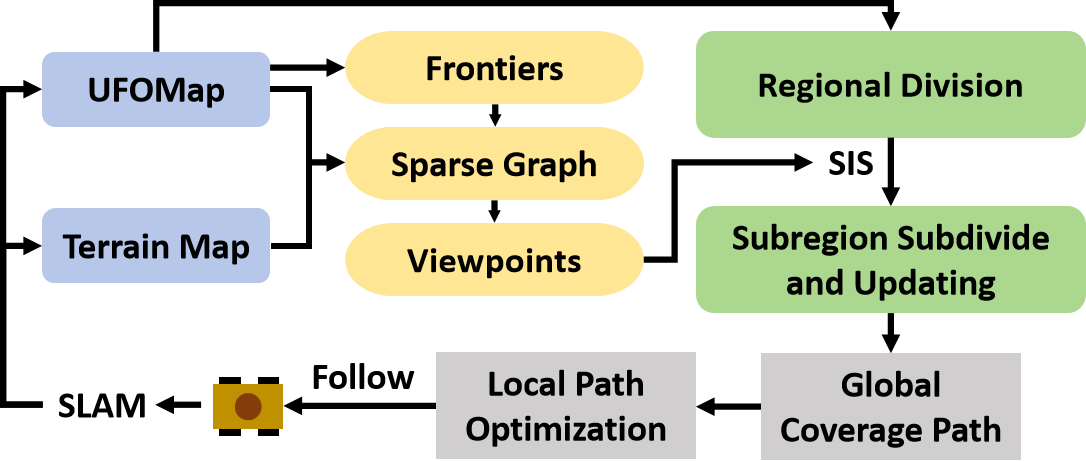}
	\caption{System overview. The environmental preprocessing is first conducted, which includes frontier detection, building sparse graph, and viewpoint generation. Then, the entire exploration environment is divided into several subregion cells, and the SIS of each cell is stored for global coverage path planning and local path optimization. Finally, the robot follows the local path to explore.}
\label{overview}
\end{figure}

\section{System Overview. }

The overview of the proposed exploration framework is presented in Fig. \ref{overview}. The exploration task is decomposed into several sub-tasks, which are then organized into a timeline for sequential execution. The task starts with rapid environment preprocessing (Sec.\ref{rapid processing}), which includes frontier extraction, sparse graph construction, and viewpoint generation. The unknown space is then divided into hierarchical subregion cells with recorded environmental information (Sec.\ref{subregion division}). Using these cells, a global coverage path is generated, guiding the refinement of the local path to cover each viewpoint efficiently (Sec.\ref{Hierarchical Planning}). The exploration ends when no viewpoints remain.

\section{Rapid Environmental Preprocessing} \label{rapid processing}
This section introduces a real-time and scalable preprocessing method for large-scale scenarios to support efficient exploration planning. It covers three main components: environment modeling and frontier detection, sparse graph construction, and viewpoint generation. 

\subsection{Mapping and Frontier Detection}
In this work, the UFOMap \cite{ufomap} is utilized for environmental representation, for its advantages of low memory consumption and fast traversal speed. Additionally, with the use of UFOMap, it becomes convenient to fetch the boundary of the updated region in the map and perform collision checking using rapid ray tracing.

As in state-of-the-art works \cite{fuel, fael}, an incremental frontier detection method is employed in this work to avoid processing the entire map. Specifically, a set $\mathcal{C}_v$ of voxels with updated occupation values is obtained from the UFOMap, and the state of each voxel $V \in \mathcal{C}_v$ is checked. If a voxel $V$ is free ($V_{free}$) and has an adjacent voxel in the unknown state ($V_{un}$) at the same height, it is identified as a frontier voxel. During exploration, each frontier voxel in the map update region $\mathcal{B}_u$ is checked to verify its validity, and invalid frontiers are removed. This method requires examining only a small subset of voxels, significantly reducing the workload for frontier extraction.

\subsection{Dynamic Sparse Graph}
Given that path search operations are time-consuming in the 3D voxel map, a dynamic sparse graph \( \mathcal{G} = (N, E) \) is maintained as a sparse environment representation, where \( N \) denotes nodes and \( E \) represents undirected weighted edges. Candidate nodes \( N_c \) are generated via biased random sampling based on the terrain map \( \mathcal{A} \), and valid nodes are added to \( \mathcal{G} \) after collision checking. Edge weights are set as the Euclidean distance between nodes, and all nodes are stored in an R*-tree\cite{rtree} to enhance computational efficiency.

\begin{equation}
\begin{cases}
   \textit{$N_c$ is in Accessible}(\mathcal{A}) \\
   \gamma < D_E(N_{\text{nearest}}, N_c) < \delta \\
\end{cases}
\label{node add}
\end{equation}
where ${Accessible}(\mathcal{A})$ represents the accessible region in $\mathcal{A}$, ensuring all nodes are reachable by the robot. The parameters $\delta$ and $\gamma$ define the minimum and maximum allowable distances between $N_c$ and the nearest node $N_{\text{nearest}}$ in $\mathcal{G}$, preventing the graph from becoming overly dense or sparse while maintaining shortest paths between nodes.

Since previously passable paths can be blocked by dynamic obstacles such as falling rocks, failing to update the sparse graph may lead to incorrect paths or inaccurate traversal distances, negatively impacting exploration efficiency and safety. To address this, the sparse graph is incrementally maintained within the updated region $\mathcal{B}_u$. Specifically, nodes and edges within $\mathcal{B}_u$ are subject to collision detection, with invalid edges caused by environmental changes being removed and valid nodes and edges being added. This method improves graph update efficiency and significantly reduces the computational cost of global graph rebuilding, enabling more efficient and robust exploration in dynamic environments.

\begin{figure}[!t]\centering
\includegraphics[width=5.7cm]{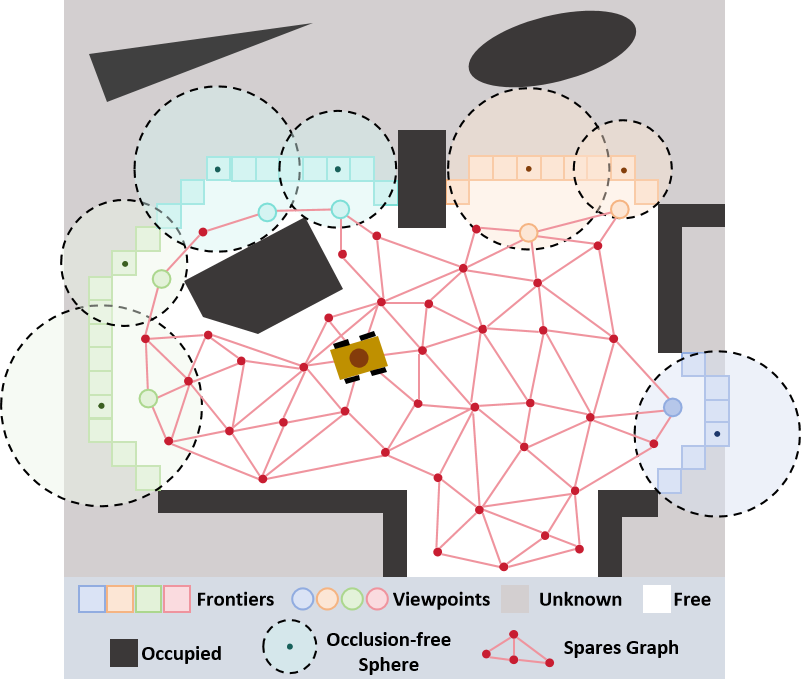}
	\caption{Visualization effect of environmental preprocessing. During the exploration process, new nodes are continuously sampled to expand the sparse graph $\mathcal{G}$. Viewpoints are obtained by sampling from $\mathcal{G}$ through generating collision-free spheres. All frontier points within the sphere can be observed from the viewpoint.}
\label{graph}
\end{figure}

\subsection{Viewpoint Generation}
In most approaches \cite{nbvp, dsvp, fuel}, the viewpoint is utilized to observe unknown voxels within the visible range of sensors, thus continuously extending the frontier. However, this process requires frequent and expensive ray tracing to assess the visibility of the unknown voxels from the viewpoint. Inspired by \cite{bubblepl, bubbleex}, we employ the collision-free sphere to generate high-quality viewpoints to observe the unknown environment. This approach aims to reduce the number of viewpoints, thus simplifying the complexity of subsequent exploration planning. The center $\mathbf{p}_s$ of the collision-free sphere is located at the frontier voxel, with its radius defined as:

\begin{equation}
    r = \min(\|\mathbf{p}_s - \mathbf{p}_o\|_2, \frac{r_{\text{sensor}}}{2})
\end{equation}
where $\mathbf{p}_o$ is the position of the nearest occupied voxel, and $r_{\text{sensor}}$ is the maximum scanning radius of the LiDAR sensor. Due to the absence of obstacles within the sphere and its convex nature, any frontier voxel in the sphere can be observed from a viewpoint inside. Thus, performing expensive ray tracing on every frontier voxel to determine its visibility from the viewpoint is unnecessary.  
Then, the node $N$ located inside the sphere and closest to its center is selected as the viewpoint. This search process is achieved by K-Nearest Neighbor search in $\mathcal{G}$. Fig. \ref{graph} shows the dynamic sparse graph and the viewpoints generated based on collision-free spheres. 

\section{Subregion-based Regional Division} \label{subregion division}
Instead of directly determining the visit order of viewpoints or frontiers, we employ subregion cells as planning units for global coverage. Each cell, managed through online decomposition and updates, enhances its ability to adapt to the scale and structural changes of unknown regions with multi-resolution.

\begin{figure}[!t]\centering
\includegraphics[width=5cm]{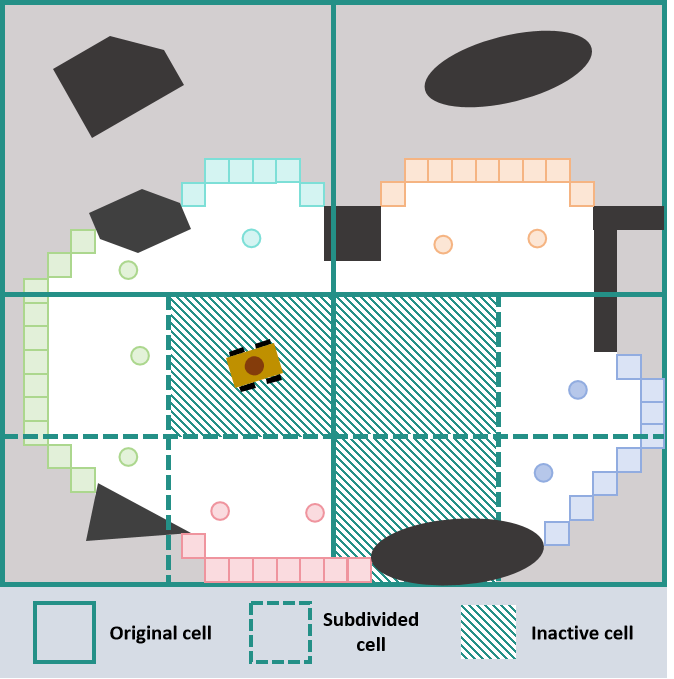}
	\caption{An example of the regional division with two levels. The original partitioned regions will be further subdivided if the known voxel ratio inside exceeds a threshold. Regions with no viewpoints inside will be considered as being in inactive status.}
\label{division}
\end{figure}

\subsection{Subregion hierarchical structure} \label{hierarchical structure}
 We utilize a quadtree-like form to continuously subdivide the whole environment into multiple subregions, characterizing the unknown regions through different levels of detail. 
 Similar to the octree's structural division, a subregion cell at level $l$ can be recursively subdivided into 2 × 2 subcells in level $l + 1$. This method aids in representing the area to be explored with varying resolutions. Fig. \ref{division} shows the hierarchical structure of the subregion cell.

At the beginning of the exploration, the total exploration space $\mathcal{O}$ is discretized into $S_o = S_{o,x} \times S_{o,y}$ cells along the axis-aligned bounding box (AABB) of the map. For any candidate subregion cell $S_i \in \left\{S_o\right\}$, as the exploration process proceeds, the volume of the unknown regions in $S_i$ gradually decreases. As a consequence, the original divided cell fails to accurately represent the unknown regions. Thus, the $S_{i}$ is further subdivided into four subcells to describe the unknown regions at a smaller resolution. The recursive subdivision method for subregions can describe unknown areas at low cost and decompose online as the exploration process progresses. This provides sufficient flexibility for representing unknown regions. Fig. \ref{division} depicts an example of the regional division.

Note that the subregion cells are only rough representations of unknown regions, which are used as basic exploration units to achieve global coverage. If the minimum area of a subregion cell is too small, a large number of subregion cells will be formed during the exploration process. This increases the complexity of the subsequent global path solving. Let $W_{\text{min}}$ and $W_{\text{max}}$ denote the minimum and maximum side lengths of the subregion cell, respectively. During the exploration, \( W_{\text{min}} \) and \( W_{\text{max}} \) should be set in an appropriate range, and the maximum level \( l_{\text{max}} \) can be calculated as:
\begin{equation}
 l_{\text{max}} = \left\lceil \log_2 \left( \frac{W_{\text{max}}}{W_{\text{min}}} \right) + 1 \right\rceil 
\end{equation}
where \( \lceil \cdot \rceil \) denotes the ceiling function.

\subsection{Subregion Information Structure}
The Subregion Information Structure \( \mathbf{SI_i} \) is composed of elemental information contained in each subregion cell \( S_i \). The hierarchy level \( l_i \) and axis-aligned bounding box $\mathbf{B}_i$ are recorded when initializing \( S_i \). The proportion of known voxels \( \mathbf{R}_{\text{known},i} \) and the viewpoint average position \( \mathbf{C}_{\text{avg},i} \) of \( S_i \) are also computed to locate the area to be explored. To serve subsequent global coverage path planning, the traveling cost \( \mathbf{L}_{\text{cost},i} \) between \( S_i \) and other subregions is also computed. The viewpoints \( \mathbf{VP}_i \) within \( S_i \) are stored in \( \mathbf{SI_i} \) for subsequent local path optimization. In addition, each subregion cell holds a status $stat_i$ between \textit{active} and \textit{inactive}. Subregion cells containing at least one viewpoint are categorized as active, while those without viewpoints are considered inactive. Only subregions with active status are eligible for inclusion in the global coverage path planning process. Table. \ref{sis} presents the meaning of each data information in SIS.

\begin{table}[h]
\centering
\caption{Information contained by SIS $\mathbf{SI_i}$ of subregion cell $S_i$.}
\resizebox{\linewidth}{!}{
\begin{tabular}{c|c}
\hline
\hline
\textbf{Information} & \textbf{Note} \\ 
\hline
$l_i$ & Hierarchical level of the cell \\
$\mathbf{B}_i$ & Axis-aligned bounding box of $S_i$ \\ 
$\mathbf{R}_{\text{known},i}$ & Proportion of known voxels in the subregion \\ 
$\mathbf{C}_{\text{avg},i}$ & Average position of viewpoints within $S_i$\\ 
$\mathbf{L}_{\text{cost},i}$ & Traveling cost from $\mathbf{C}_{\text{avg},i}$ of $S_i$ to other cells\\ 
$\mathbf{VP}_i$ & Viewpoints in subregion \\ 
$stat_i$ & \textit{active} or \textit{inactive} status of $S_i$ \\
\hline
\hline
\end{tabular}
}

\label{sis}
\end{table}

\subsection{Online Subregion Updating} \label{Online Subregion Updating}
We maintain a set $\mathcal{V}$ that stores all divided subregions. Initially, the entire unknown space $\mathcal{O}$ is divided into $S_o = S_{o,x} \times S_{o,y}$ cells at level $l = 1$. During exploration, as the map updates, the subregion information $\mathbf{SI_i}$ of $S_i$ is modified accordingly. Therefore, all subregion cells intersecting with the map update region $\mathcal{B}_u$ need to recompute their SIS. Algorithm \ref{alg: Online Updating} outlines the online subregion cell updating process. First, traverse the set \( \mathcal {V} \) to filter the cell \( S_i \) that overlaps with $\mathcal{B}_u$ (Line 3). Then, some attributes such as the proportion of known voxels \( \mathbf{R}_{\text{known},i} \) and the status $ stat_i$ is updated (Line 4). For $S_i$ in active status, if $\mathbf{R}_{\text{known},i}$ exceeds a threshold $\sigma_k$ and $l_i < l_{\text{max}}$, the cell is subdivided, removed from $\mathcal{V}$, and its subcells are added after recalculating their SIS (Lines~5--12). Otherwise, passable paths from $S_i$ to other subregions are updated using the graph search in $\mathcal{G}$ (Line~15). Since the traveling cost $\mathbf{L}_{\text{cost},i,j}$ is symmetric, it is updated for both $i \to j$ and $j \to i$.

\begin{algorithm}[!t]
    \caption{Online Subregion Updating}
    \label{alg: Online Updating}
    Initialize $\mathcal{V} = \{S_o\}$ \;
    \For{$S_i \in \mathcal{V}$}{
        \If{\texttt{overlapCheck}($S_i, \mathcal{B}_u$)}{
            \texttt{updateAttributes}($S_i$) \;
            \If{\texttt{isActive}($S_i$)}{
                \If{$\mathbf{R}_{\text{known},i} > \sigma_k$ \textbf{and} $l_i < l_{\text{max}}$}{
                    $S_{\text{new}} \gets \texttt{splitRegion}(S_i)$ \;
                    $\mathcal{V} \gets \mathcal{V} \setminus \{S_i\}$ \;
                    \For{$S_j \in S_{\text{new}}$}{
                        \texttt{computeSIS}($S_j$) \;
                        $\mathcal{V} \gets \mathcal{V} \cup \{S_j\}$ \;
                    }
                }
                \Else{
                    \texttt{updateTravelCost}($S_i, \mathcal{V}$) \;
                }
            }
        }
    }
\end{algorithm}

\section{Hierarchical Planning Strategy} \label{Hierarchical Planning}
\begin{figure}[!t]\centering
\includegraphics[width=6cm]{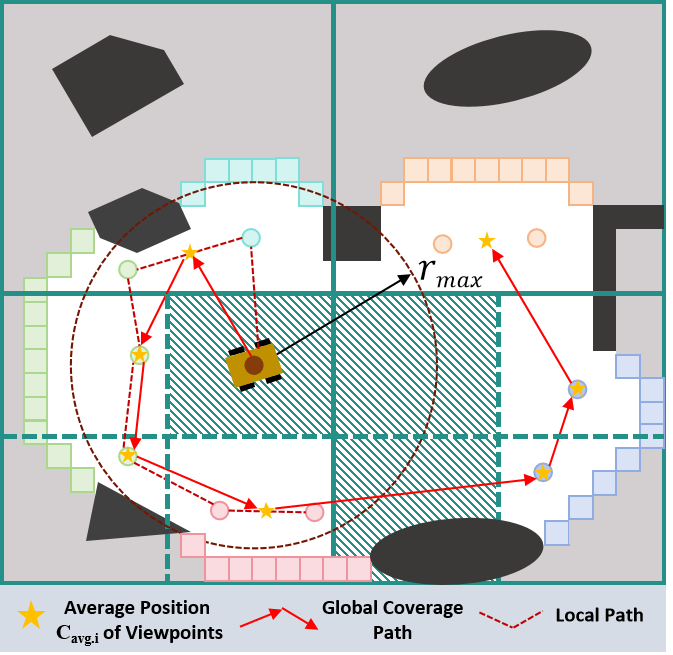}
	\caption{Illustration of hierarchical planning. The red arrows represent the global coverage path $\mathcal{G}_{\text{global}}$ that sequentially enters all subregions in active status, while the dashed lines indicate the optimized local path $\mathcal{T}_{\text{local}}$ guided by the global path $\mathcal{G}_{\text{global}}$. The local path visits all viewpoints within $r_{\text{max}}$.}
\label{planning}
\end{figure}

At this stage, it is imperative to consider organizing the access sequence of these subregions to get an efficient global coverage path. Then, guided by the global path, the visit order of local viewpoints is refined. As depicted in Fig. \ref{planning}, we employ a hierarchical planning paradigm to make a coarse-to-fine decision on the next visit goal.

\subsection{Global Coverage Path Planning}
The objective of global coverage path planning is to find a global path $\mathcal{G}_{\text{global}}$ that enters each subregion cell $S_i$ and efficiently covers the entire unknown environment. Similar to \cite{fuel}, we also formalize it as a variant of the TSP. Assuming there are \(N_{\text{act}}\) subregions in active status, solving this TSP requires computing a \( (N_{\text{act}}+1) \)-dimensional distance cost matrix $\mathbf{M}_{\text{cost}}$, as the current robot pose $\mathbf{p}_0$ also needs to be considered. Since the traveling cost $\mathbf{L}_{\text{cost},i,j}$ of each subregion has already been computed in Sec.\ref{Online Subregion Updating}, \(N_{\text{act}} \times N_{\text{act}}\) blocks can be directly filled in:
\begin{equation}
    \mathbf{M}_{\text{cost}} (S_i, S_j) = \mathbf{L}_{\text{cost},i,j}, i,j \in \left\{1,2,\cdots,N_{\text{act}}\right\}
    \label{1}
\end{equation}
Note that \( \mathbf{L}_{\text{cost},i, j} \) is computed based on the feasible path between $\mathbf{C}_{\text{avg},i}$ of \( S_i \) and  $\mathbf{C}_{\text{avg},j}$ of \( S_j \), which is obtained in  Sec.\ref{Online Subregion Updating} by executing the A* algorithm in the sparse graph $\mathcal{G}$. And only \(N_{\text{act}}\) additional calculations of the distance $\mathbf{L}_{\text{cost},\mathbf{p}_0,k}$ from the robot pose $\mathbf{p}_0$ to each subregion cell are required:
\begin{equation}
    \mathbf{M}_{\text{cost}} (0, S_k) = \mathbf{L}_{\text{cost},\mathbf{p}_0,k}, k \in \left\{1,2,\cdots,N_{\text{act}}\right\}
    \label{2}
\end{equation}
Also adding a zero distance cost in the first column of $\mathbf{M}_{\text{cost}}$ simplifies the TSP problem into an asymmetric TSP:
\begin{equation}
    \mathbf{M}_{\text{cost}} (S_k, 0) = 0, k \in \left\{1,2,\cdots,N_{\text{act}}\right\}
    \label{3}
\end{equation}

Thus far, the cost matrix $\mathbf{M}_{\text{cost}}$ is constructed, and the TSP is then solved using a Lin-Kernighan-Helsgaun heuristic \cite{LKH}. The output of the solver is the sequence of accessing \(N_{\text{act}}\) subregion cells $\mathcal{G}^*_{\text{global}}=[S_1,S_2,\cdots,S_{N_{\text{act}}}]$, where $\mathcal{G}^*_{\text{global}}$ denotes the optimal access sequence of the subregion cells. This access sequence represents the shortest path that can sequentially pass through all active subregions.

\subsection{Local Path Optimization} 
The global coverage path identifies an efficient entry sequence for all active subregion cells. However, there are still numerous viewpoints within each cell that require visits to expand the frontiers. At this stage, our focus remains on the cumulative traveling distance to visit these viewpoints, rather than using a greedy strategy to evaluate their revenue.

To this end, we need to find the shortest visiting path $\mathcal{T}^*_{\text{local}}$ that connects viewpoints while also considering the endpoint constraints under the global coverage path $\mathcal{G}^*_{\text{global}}$. Given the visiting order of subregion cells $\mathcal{G}^*_{\text{global}}=[S_1,S_2,\cdots,S_{N_{\text{act}}}]$, the local path is denote as:
\begin{equation}
    \mathcal{T}_{\text{local}} = [\mathbf{v}_1^1, \mathbf{v}_1^2, \cdots][\mathbf{v}_2^1, \mathbf{v}_2^2, \cdots] \ldots [\mathbf{v}^1_{N_{\text{act}}}, \mathbf{v}^2_{N_{\text{act}}}, \cdots] 
    \label{7}
\end{equation}
where \([ \cdot ]\) represents a local path segment in a subregion cell, and $\mathbf{v}_i^j \in \mathbf{VP}_i$, which denotes the \(j\)-th viewpoint in the \(i\)-th subregion. This formulation ensures that under the constraints of $\mathcal{G}^*_{\text{\text{global}}}$, viewpoints in \(S_i\) are visited before those in \(S_{i+1}\).

The above problem can be viewed as a TSP with endpoint constraints. Meanwhile, guided by the global path, it is unnecessary to consider viewpoints that are distant from the robot. Only viewpoints within a radius $r_{\text{max}}$ are considered. Similar to the procedures outlined in Eq. \ref{1} and Eq. \ref{3}, the A* algorithm is used in $\mathcal{G}$ to search for feasible paths between viewpoints to calculate distance cost $\mathbf{D}_{\text{cost}}$, and zero costs are added to form the asymmetric TSP. The only difference is that in addition to the distance cost $\mathbf{D}_{\text{cost},\mathbf{p}_0}$ from the robot pose $\mathbf{p}_0$ to each viewpoint, a motion consistency cost $c_c$ is also incorporated to discourage the back-and-forth behavior:
\begin{equation}
\begin{aligned}
    \mathbf{M}_{\text{cost}} (0, \mathbf{v}_l) = \mathbf{D}_{\text{cost},\mathbf{p}_0,l} + \mu_c \cdot c_c(\mathbf{p}_0, \mathbf{v}_l) &, \\
    l \in \left\{1,2,\cdots,N_{\mathbf{v}}\right\}&
\end{aligned}
\end{equation}

\begin{equation}
    c_{c}(\mathbf{p}_0, \mathbf{v}_l) = \cos^{-1} \frac{(\mathbf{p}_{\mathbf{v}_{l}} - \mathbf{p}_0) \cdot \mathbf{v}_0}{\|\mathbf{p}_{\mathbf{v}_{l}} - \mathbf{p}_0\| \|\mathbf{v}_0\|} 
\end{equation}
where $\mathbf{v}_l$ represents the $l$-th viewpoint among a total of $N_{\mathbf{v}}$ viewpoints in range $r_{\text{max}}$, $\mu_c$ is the tuning factor, $\mathbf{v}_0$ is the current velocity of the robot, and $\mathbf{p}_{\mathbf{v}_l}$ represents the position of the viewpoint $\mathbf{v}_l$. 

Note that the optimization problem mentioned above is not merely finding the shortest path visiting each viewpoint. As stated in Eq. \ref{7}, it must satisfy the endpoint constraint, meaning $\mathbf{v}_i^j$ must be visited before $\mathbf{v}^1_{i+1}$. This constraint is included before solving with the TSP solver.

After obtaining $\mathcal{T}^*_{\text{local}}$, the resulting discrete waypoints consist of adjacent viewpoints. It is crucial to connect these waypoints using the shortest passable path between viewpoints, which is then transmitted to the motion planning module as the executable path to guide the robot's movement. The passable path between viewpoints has already been determined during the calculation of $\mathbf{D}_{\text{cost}}$, ensuring that this process incurs no additional computational overhead.

\section{Experiments and Analysis}

\begin{figure}[!t]
\centering
\subfigure[Scene 1]{
    \includegraphics[width=2.5cm]{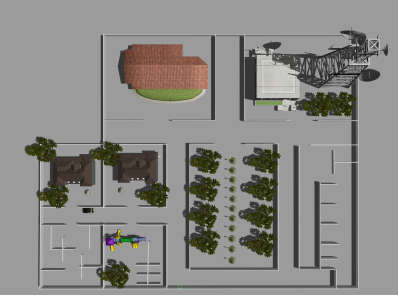}}
\subfigure[Scene 2]{
    \includegraphics[width=2.5cm]{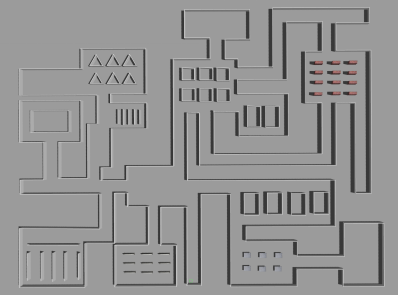}}
\subfigure[Scene 3]{
    \includegraphics[width=2.5cm]{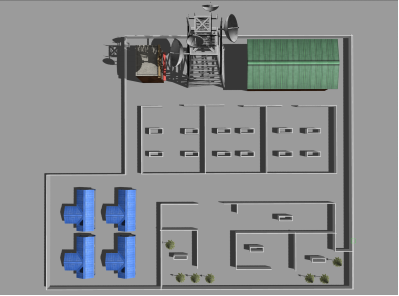}}
\caption{The three simulation environments built in the Gazebo simulator.}
\label{scenes}
\end{figure}

\subsection{Simulation Experiment}
 The simulation experiments are conducted on an Intel Core i5-11320H@3.20 GHz CPU with Ubuntu 20.04 LTS and 16 GB RAM. We evaluate the performance of our method in simulation environments by comparing it with three state-of-the-art methods, e.g., TARE\cite{tare}, DSVP\cite{dsvp}, and GBP \cite{gbp}.
The compared metrics include exploration volume, traveling length, and algorithm runtime. Each method is run 10 times in environments depicted in Fig. \ref{scenes}. All tests utilize the exploration algorithm as the top-level decision module, while the middle layers such as terrain analysis and obstacle avoidance employ an open-source framework \cite{framework} from the Robotics Institute at Carnegie Mellon University. For all approaches, the starting position is consistent, with a limited lidar detection radius of 13 m and a maximum robot speed set to 2.0 m/s.

\begin{table}[h]
\caption{Simulation Results in Three Environments}
\begin{center}
\resizebox{8.5cm}{!}{
\begin{tabular}{c|c|c|c|c|c|c}
\hline
\hline
    \multirow{2}{*}{\textbf{Scene}} & \multirow{2}{*}{\textbf{Method}} & \multicolumn{2}{c|}{\textbf{Distance}(\textit{m})} & \multicolumn{2}{c|}{\textbf{Time Duration}(\textit{s})}  & \multirow{2}{*}{\makecell[c]{\textbf{Success} \\ \textbf{Number}}} \\
\cline{3-6}

&  & \textit{Avg} & \textit{Std} & \textit{Avg} & \textit{Std} \\
\hline
\multirow{4}{*}{\makecell[c]{Scene 1\\ ($100 \times 80$)}} 
& GBP  & $>$ 824.21 & - & $>$ 1200 & - & 0\\
& DSVP & 1440.85 & 67.18  & 866.27 & 30.27 & 10\\
& TARE & 1348.60 & 52.11  & 822.78 & 33.41  & 10\\
& \textbf{Ours}  & \textbf{992.62} & \textbf{10.13} & \textbf{589.60}& \textbf{5.25}  & 10\\
\hline

\multirow{4}{*}{\makecell[c]{Scene 2\\ ($137 \times 104$)}} 
& GBP  & $>$ 1235.55 & -  & $>$ 1600 & -  & 0\\
& DSVP & 1697.53 & 83.58  & 1057.24 & 61.34  & 10\\
& TARE & 1568.47 & 60.38  & 885.91 & 29.93  & 10\\
& \textbf{Ours}  & \textbf{1439.82} & \textbf{29.39} & \textbf{819.56} & \textbf{13.22}  & 10\\
\hline

\multirow{4}{*}{\makecell[c]{Scene 3\\ ($80 \times 80$)}} 
& GBP  & $>$ 606.27 & -  & $>$ 800 & -  & 0\\
& DSVP & 845.43 & \textbf{18.78}  & 514.02 & \textbf{13.49}  & 10\\
& TARE & 778.55 & 46.53   & 492.99 & 28.46  & 10\\
& \textbf{Ours}  & \textbf{630.76} & 27.13 & \textbf{390.40} & 20.10  & 10\\
\hline
\hline
\end{tabular}
}
\label{experiment}
\end{center}
\end{table}

\begin{figure*}[htbp]
\centering
    \subfigure[Scene 1]{
    \includegraphics[width=5cm]{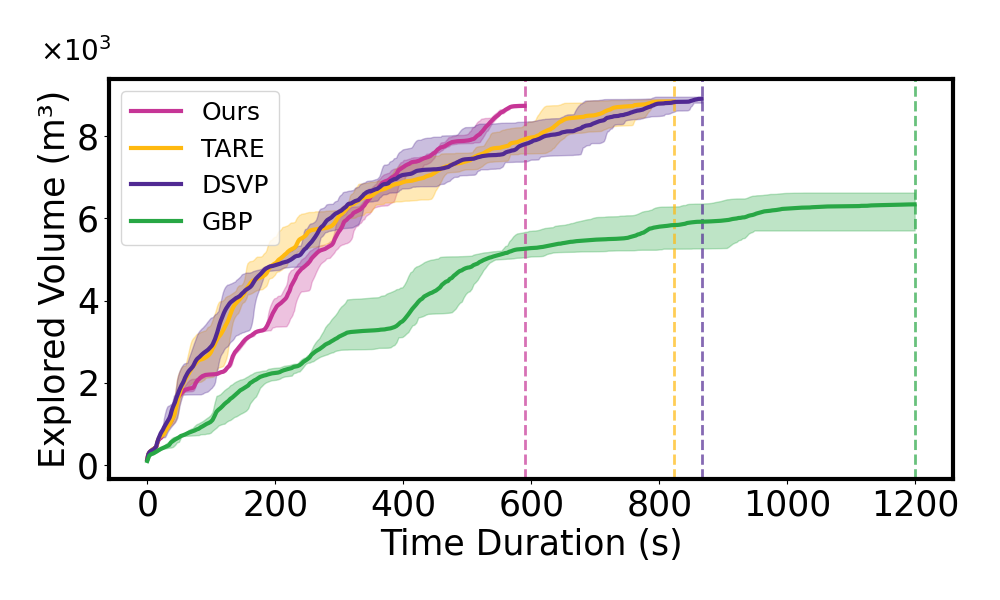}}
\subfigure[Scene 2]{
    \includegraphics[width=5cm]{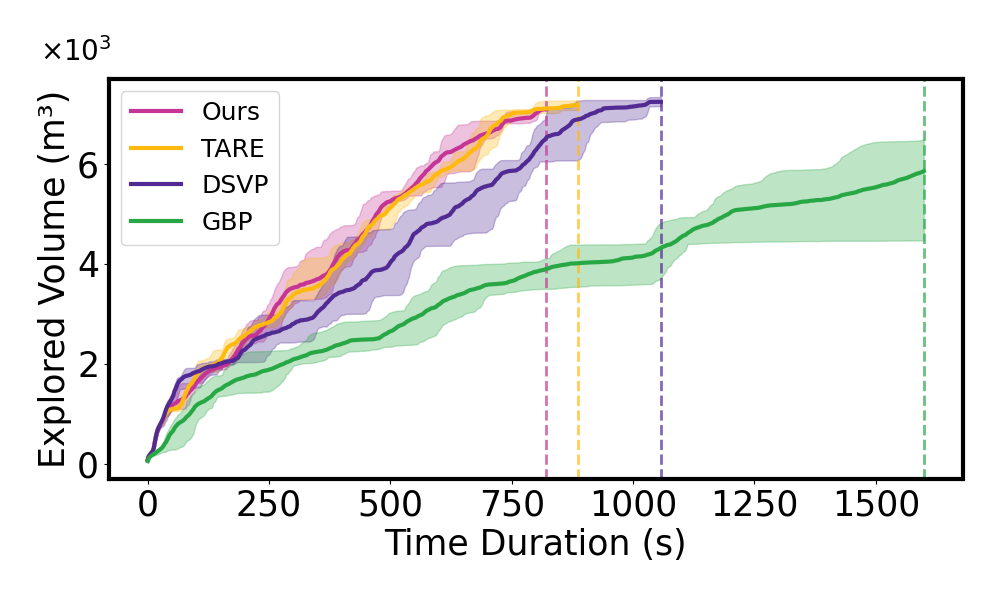}}
\subfigure[Scene 3]{
    \includegraphics[width=5cm]{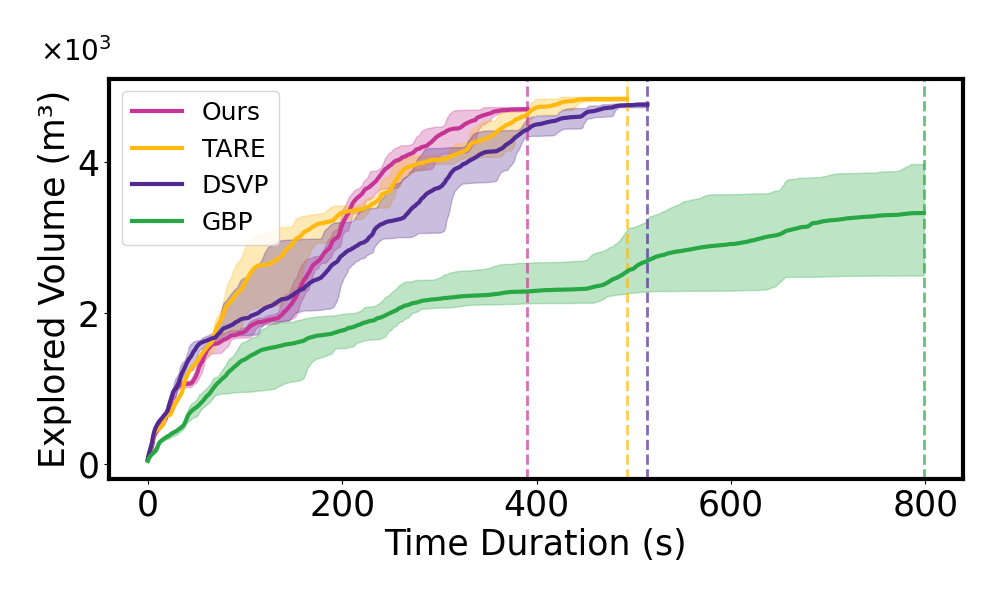}}
\caption{The exploration progress of four methods in each scene. The solid line represents the average exploration progress, while the shading region in color is formed by the upper-bound and lower-bound of the exploration progress. The vertical dashed line represents the average exploration end time.}
\label{explored}
\end{figure*}

\textit{1) Exploration Performance}: Table. \ref{experiment} presents the specific experimental results and statistics, while Fig. \ref{explored} visualizes the exploration progress curves for each method. The exploration time limit for each scenario is set to twice the maximum exploration time of our method. All methods except GBP cover the entire exploration space. However, our proposed method consistently outperforms the others in each environment. In the middle and late stages of exploration, GBP struggles to relocate and switch exploration goals in open areas, causing the robot to freeze in place and preventing it from fully exploring the entire environment. DSVP employs a greedy exploration strategy, selecting branches with the highest gain on the expanded tree. This often leads to local optima and back-and-forth movements due to its focus on high-gain regions while ignoring lower-gain areas.

Our approach, similar to TARE, addresses the TSP of arranging the visitation order in the unknown region. However, we transform the coverage planning problem into an open-loop asymmetric TSP, eliminating the need to return to the starting point after each iteration. This approach avoids the necessity of considering the return path from the last node to the start, which is more typical in exploration tasks. In contrast, TARE addresses a closed-loop TSP, where some regions are scheduled to be visited on the return trip. Since exploration tasks generally do not require a return to the starting point, the closed-loop TSP can result in an inefficient order of access, leading to missed areas and subsequent backtracking. By employing an open-loop asymmetric TSP, our strategy circumvents these issues, thereby enhancing exploration efficiency. Moreover, in our method, neither the global coverage path nor the local path accounts for information gain. We believe that introducing other gains when constructing the cost matrix of the TSP may interfere with the solving process and lead to suboptimal solutions. Since exploration is a long-term task, minimizing exploration distance should be a primary goal.

This does not mean that information gain is trivial, but rather that once sufficient information gain is achieved, its quantity can be disregarded. In our method, subregion cells without viewpoints are excluded from global coverage path planning because they lack sufficient information gain. Similarly, all viewpoints are based on the collision-free sphere and can see frontiers with ample information gain. From this perspective, our calculation of information gain is implicitly included in the environment preprocessing process.

\begin{figure*}[htbp]
\centering
    \subfigure[Scene 1]{
    \includegraphics[width=5cm]{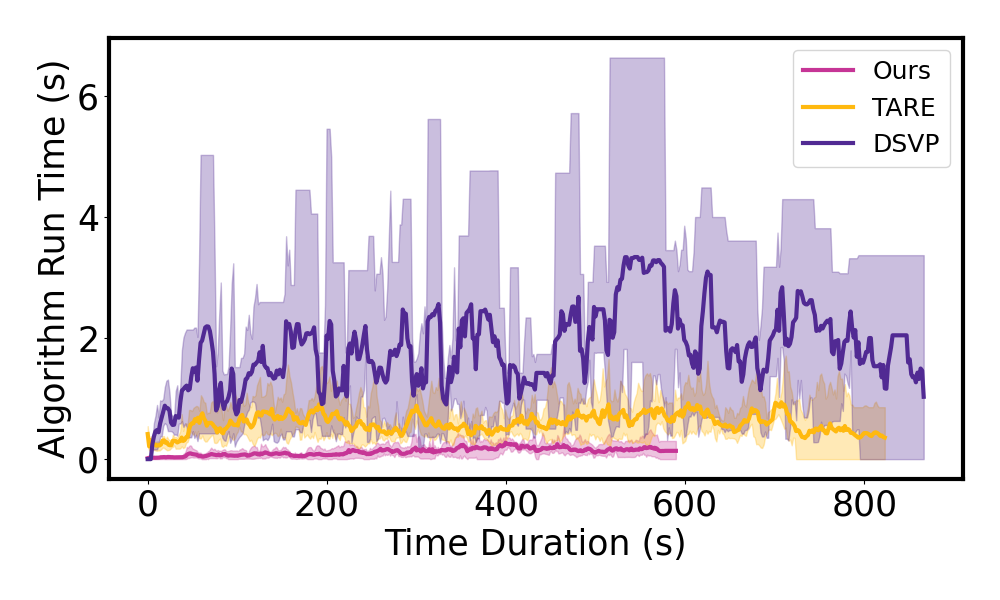}}
\subfigure[Scene 2]{
    \includegraphics[width=5cm]{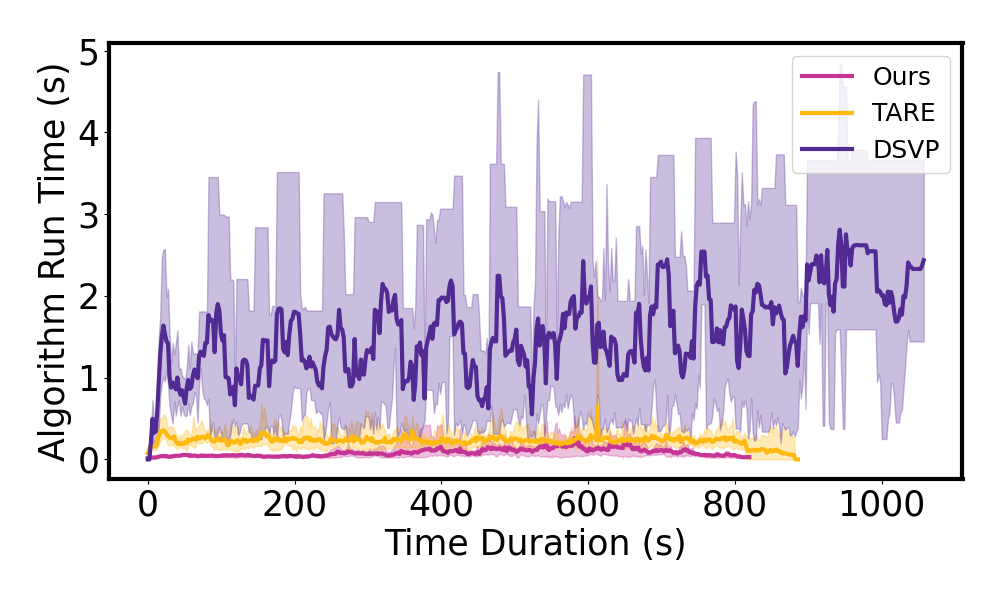}}
\subfigure[Scene 3]{
    \includegraphics[width=5cm]{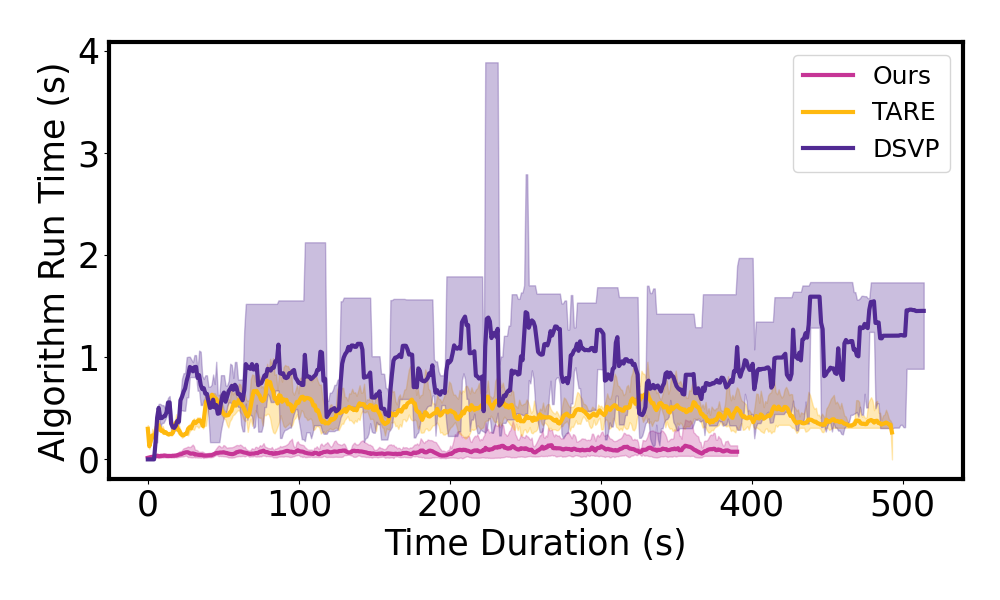}}
\caption{The algorithm runtime of three methods in each scene. The solid line represents the average runtime. while the shading region in color is formed by the upper-bound and lower-bound of the runtime.}
\label{runtime}
\end{figure*}

\textit{2) Computational Efficiency}: We also conduct a comprehensive analysis with DSVP and TARE regarding algorithm runtime to highlight the advantages of our proposed method, as shown in Fig. \ref{runtime}. Compared to TARE and DSVP, our method exhibits the shortest runtime due to its rapid environmental preprocessing and hierarchical planning strategy. DSVP suffers from long iteration times, leading to delayed decision-making, particularly in large-scale environments. The computational burden of DSVP primarily lies in evaluating the gain of branches, which involves counting many unknown voxels and using ray tracing to check sensor visibility. Consequently, more than 90\% of the runtime is consumed by this raycast-based gain computation. In contrast, our approach samples viewpoints from nodes on the sparse graph using the collision-free sphere, leveraging the number of frontier voxels observable from each viewpoint as an information gain metric. This process requires minimal ray tracing to assess visibility from the spherical region to nodes.

TARE addresses the access order of local planning horizons and global subspaces. However, it generates a large number of viewpoints and subspaces during exploration, especially in outdoor areas of Scene 1 and Scene 3. These numerous viewpoints and subspaces need to be accessed sequentially by solving the TSP, increasing the complexity of the solution. Our method continuously decomposes the subregion to represent the unknown region with appropriate resolution. All viewpoints are generated by collision-free spheres, minimizing redundant viewpoints. Guided by the global-level coverage path, we solve the visiting sequence for a small number of viewpoints, thereby reducing the complexity of TSP solving.

\subsection{Ablation Experiment}
As outlined in Sec.\ref{hierarchical structure}, a variable-resolution mechanism is employed to enhance adaptability to the scale and structural changes of the unknown region. However, how do different resolution settings specifically affect exploration efficiency? Moreover, to what extent do these settings influence the computational requirements of the proposed approach?

We design an experimental variation using two fixed resolutions ($W=8 m$ and $W=25 m$) for subregion division. The experimental results, as shown in Table. \ref{ablation}, indicate that subregions with finer resolution achieve better exploration performance. This result aligns with intuition, as higher-resolution representations provide more precise global coverage paths. However, this also significantly increases the complexity of solving the TSP, leading to higher computation times, which deviates from the intended efficiency of our coarse-to-fine planning framework. Conversely, overly coarse resolutions fail to provide effective global exploration guidance, thereby slowing down the overall exploration process.

In contrast, the hierarchical subregion structure dynamically adjusts the spatial size of subregions based on environmental characteristics, enabling flexible representation of unknown areas. Finer subdivisions are applied to large covered regions to enhance exploration precision, while lower-resolution representations are employed for vast uncovered regions to reduce computational overhead. This dynamic adjustment strategy not only accurately captures the features of unexplored areas but also significantly reduces the total number of subregions, achieving a desirable balance between exploration efficiency and computational cost.

\begin{table}[h]
\caption{Impact of Subregion Resolutions on Exploration Performance}
\begin{center}
\resizebox{8.5cm}{!}{
\begin{tabular}{c|c|c|c|c|c|c|c}
\hline
\hline
    \multirow{2}{*}{\textbf{Scene}} & \multirow{2}{*}{\textbf{Resolution}} & \multicolumn{2}{c|}{\textbf{Distance}(\textit{m})} & \multicolumn{2}{c|}{\textbf{Time Duration}(\textit{s})} & \multicolumn{2}{c}{\textbf{Run Time}(\textit{s})} \\
\cline{3-8}

&  & \textit{Avg} & \textit{Std} & \textit{Avg} & \textit{Std} & \textit{Avg} & \textit{Std}\\
\hline
\multirow{4}{*}{\makecell[c]{Scene 1\\ ($100 \times 80$)}} 
& $W=8 m$ & 974.34 & 14.37  & 587.27 & 8.87 & 0.245 & 0.124 \\
& $W=25 m$ & 1121.70 & 15.32  & 655.88 & 10.41 & 0.096 & 0.068 \\
& \cellcolor{gray!20}{$Variable$}  & \cellcolor{gray!20}{992.62} & \cellcolor{gray!20}{10.13} & \cellcolor{gray!20}{589.60}& \cellcolor{gray!20}{5.25}  & \cellcolor{gray!20}0.125 & \cellcolor{gray!20}0.094\\
\hline

\multirow{4}{*}{\makecell[c]{Scene 2\\ ($137 \times 104$)}} 
& $W=8 m$ & 1440.11 & 30.45  & 818.96 & 21.77 & 0.182 & 0.098 \\
& $W=25 m$ & 1597.68 & 26.98  & 915.49 & 16.41 & 0.063 & 0.045 \\
& \cellcolor{gray!20}$Variable$  & \cellcolor{gray!20}1439.82 & \cellcolor{gray!20}{29.39} & \cellcolor{gray!20}{819.56} & \cellcolor{gray!20}{13.22}  & \cellcolor{gray!20}0.085 & \cellcolor{gray!20}0.072\\
\hline

\multirow{4}{*}{\makecell[c]{Scene 3\\ ($80 \times 80$)}} 
& $W=8 m$ & 611.85 & 25.18  & 377.13 & 18.87 & 0.122 & 0.089 \\
& $W=25 m$ & 756.67 & 19.76  & 478.78 & 23.21 & 0.068 & 0.037 \\
& \cellcolor{gray!20}$Variable$  & \cellcolor{gray!20}630.76 & \cellcolor{gray!20}27.13 & \cellcolor{gray!20}390.40 & \cellcolor{gray!20}20.10  & \cellcolor{gray!20}0.075 & \cellcolor{gray!20}0.043\\
\hline
\hline
\end{tabular}
}
\label{ablation}
\end{center}
\end{table}

\subsection{Real-world Experiment}
The real-world experiment is conducted in an underground garage with parked vehicles, as shown in Fig. \ref{real_map}, with the size of the garage being 50 m $\times$ 50 m. The environment includes moving vehicles and narrow passageways formed between the parked vehicles. Fig. \ref{real_map} also shows our real-world ground platform with an AMD Ryzen 7 7840HS@3.80 GHz CPU and 32 GB RAM using Ubuntu 20.04 LTS. The platform is topped with the RoboSense Hellos-16P LiDAR sensor. The Direct LiDAR-Inertial Odometry \cite{dlio} is used as the SLAM module. The max update range radius of the UFOMap is set to 8 m, and the max speed of the robot is set to 1.0 m/s.

Fig. \ref{real_map} shows the trajectory of the exploration result, and there are no redundant paths during the whole exploration process. At the beginning of the exploration, the environment is unknown to the robot. The robot explores \textbf{5673.25} m$^3$ in \textbf{175.53} s and moves \textbf{157.53} m. Our method completely explores the entire garage without leaving any areas. The experiment proves our method's efficiency and effectiveness.

\section{Conclusion}
In this paper, we present an autonomous exploration system that is capable of efficiently modeling unknown environments. The method introduces a rapid environmental preprocess approach to provide basic information for subsequent exploration planning, greatly reducing the computational load. We employ a hierarchical planning strategy to arrange the entry sequence of each subregion cell and determine the next viewpoint to be visited from coarse to fine, striking a balance between exploration efficiency and computational cost. Our method emphasizes the importance of global coverage path planning for long-term exploration tasks rather than focusing solely on immediate gains, especially in large-scale environments. This system is expected to be deployed on wheeled-legged and aerial robots in the future, enabling autonomous exploration in various complex terrains.

\addtolength{\textheight}{-12cm}  

\bibliographystyle{IEEEtran}
\bibliography{ref}
\end{document}